\documentclass[twoside,11pt]{article}

\usepackage{jmlr2e}
\usepackage{xcolor}

\usepackage{lastpage}

\newcommand{\omltversion}{OMLT 1.0}

\ShortHeadings{OMLT: Optimization \& Machine Learning Toolkit}{Ceccon, Jalving, et al.}
\firstpageno{1}

\begin{document}

\jmlrheading{23}{2022}{1-\pageref{LastPage}}{3/22; Revised 7/22}{11/22}{22-0277}{Francesco Ceccon, Jordan Jalving, Joshua Haddad, Alexander Thebelt, Calvin Tsay, Carl D Laird, Ruth Misener}

\title{OMLT: Optimization \& Machine Learning Toolkit}

\author{\name Francesco Ceccon$^{1,}$\thanks{These authors contributed equally. $^{\dagger}$. These authors contributed equally.} \email francesco.ceccon14@imperial.ac.uk
       \AND
       \name Jordan Jalving$^{2,*}$ \email jhjalvi@sandia.gov
       \addr
       \AND
       \name Joshua Haddad$^2$ \email jihadda@sandia.gov
       \AND
       \name Alexander Thebelt$^1$ \email alexander.thebelt18@imperial.ac.uk
       \AND
       \name Calvin Tsay$^1$ \email c.tsay@imperial.ac.uk
       \AND
       \name Carl D Laird$^{3, \dagger}$ \email claird@andrew.cmu.edu
       \AND
       \name Ruth Misener$^{1, \dagger}$ \email r.misener@imperial.ac.uk \\
       \addr $^1$ Department of Computing, Imperial College London, 180 Queen's Gate, SW7 2AZ, UK \\
       $^2$ Center for Computing Research, Sandia National Laboratories, Albuquerque, NM 87123, USA \\
       $^3$ Department of Chemical Engineering, Carnegie Mellon University, Pittsburgh, PA 15213, USA
       }

\editor{Sebastian Schelter}

\maketitle

\begin{abstract}
The optimization and machine learning toolkit (OMLT) is an open-source software package incorporating neural network and gradient-boosted tree surrogate models, which have been trained using machine learning, into larger optimization problems. We discuss the advances in optimization technology that made OMLT possible and show how OMLT seamlessly integrates with the algebraic modeling language Pyomo. We demonstrate how to use OMLT for solving decision-making problems in both computer science and engineering.
\end{abstract}


\section{Introduction}

The optimization and machine learning toolkit (\url{https://github.com/cog-imperial/OMLT}, \omltversion) is an open-source software package enabling optimization over high-level representations of neural networks (NNs) and gradient-boosted trees (GBTs).
Optimizing over trained surrogate models allows integration of NNs or GBTs into larger decision-making problems.
Computer science applications include maximizing a neural acquisition function \citep{volpp2019meta} or verifying neural networks \citep{botoeva2020efficient}.
In engineering, machine learning models may replace complicated constraints or act as surrogates in larger design and operations problems \citep{henao2011surrogate}.
\omltversion{} supports GBTs through an ONNX\footnote{\url{https://github.com/onnx/onnx}} interface and NNs through both ONNX and Keras \citep{chollet2015keras} interfaces. OMLT transforms these pre-trained machine learning models into the algebraic modeling language Pyomo \citep{bynum2021pyomo} to encode the optimization formulations.

Mathematical optimization solver software requires, as input, a formulation including the decision variable(s), objective(s), constraint(s), and any parameters. OMLT automates the otherwise tedious and error-prone task of translating already-trained NN and GBT models into optimization formulations suitable for solver software. For example, the ReLU NN example in \texttt{neural\_network\_formulations.ipynb} switches (in $\leq1$ line of code) between 3 formulations (\emph{complementarity}, \emph{big-M}, \emph{partition}) with 248, 308, and 428 constraints, respectively. \omltversion{} saves the time needed to code/debug each optimization formulation.

\begin{figure}
    \centering
    \includegraphics[height=1.9cm]{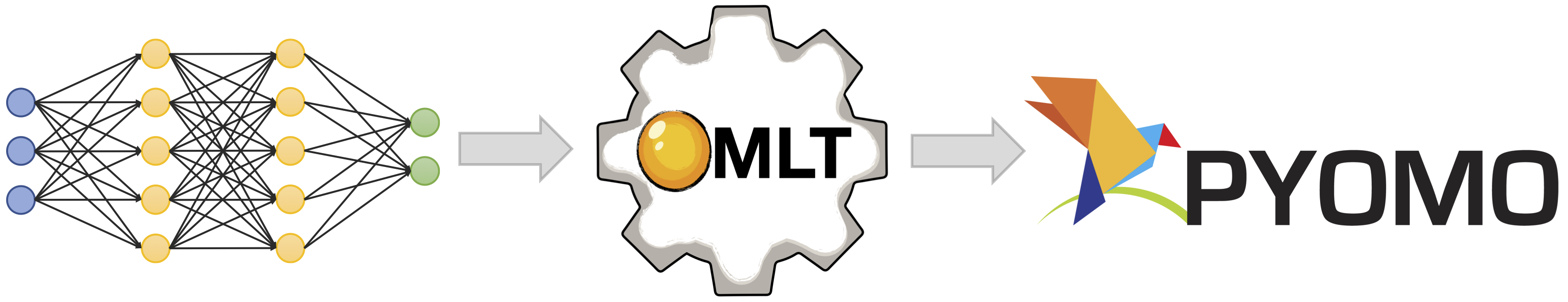}
    \caption{OMLT automatically translates high-level representations of machine learning models into variables and constraints suitable for optimization.}
    \label{fig:omlt}
\end{figure}

\section{Use cases for optimizing trained machine learning surrogates}

Complete NN verification is an AI use case for embedding trained NNs into an optimization problem \citep{tjeng2017evaluating}.
One such verification problem asks: Given a trained NN, a labeled target image, and a distance metric, does an image with a different, adversarial label exist within a fixed perturbation? Such NN verification can be formulated as an optimization problem \citep{lomuscio2017approach}. Our \texttt{mnist\_example\_\{dense, cnn\}.ipynb} notebooks verify dense and convolutional NNs on MNIST \citep{lecun2010mnist}.
The problems in the \texttt{mnist\_example\_\{dense, cnn\}.ipynb} notebooks could also be addressed effectively with dedicated NN verification tools, e.g., Beta-CROWN \citep{wang2021beta}. Related applications however, e.g., minimally distorted adversaries \citep{croce2020minimally} and lossless compression \citep{serra2020lossless}, cannot be addressed by the dedicated NN verification tools. These other applications can still be implemented in OMLT.

An engineering use case is the \texttt{auto-thermal-reformer\{-relu\}.ipynb} notebook, which develops an NN surrogate with data from a process model built using IDAES-PSE \citep{lee2021idaes}.
Here, the goal is to build surrogate models for complex processes to improve optimization convergence reliability or replace simulation-based models with equation-oriented optimization formulations. \cite{KILWEIN2021919}  mix optimal power flow constraints with security constraints learned from data.
Other examples include grey-box optimization or hybrid mechanistic / data-driven optimization \citep{BOUKOUVALA2016701,boukouvala2017global,wilson2017alamo,boukouvala2017argonaut,huster2020deterministic,THEBELT2022117469}.

\section{Library design}\label{sec:library-design}

\textbf{Input interface.} OMLT uses ONNX as an input interface because ONNX implicitly allows OMLT to support packages such as Keras \citep{chollet2015keras}, PyTorch \citep{NEURIPS2019_9015}, and TensorFlow \citep{tensorflow2015-whitepaper} via the ONNX interoperatability features.

\medskip

\noindent
\textbf{Formulating surrogate models as a Pyomo block.} OMLT uses Pyomo, a Python-based algebraic modeling language for optimization \citep{bynum2021pyomo}. Most machine learning frameworks use Python as the primary interface, so Python is a natural base for OMLT.
Pyomo has a flexible modeling interface to Pyomo-enabled solvers: switching solvers allows OMLT users to select the best optimization solver for an application without explicitly interfacing with each solver.
OMLT heavily relies on Pyomo. First, Pyomo's efficient auto-differentiation of nonlinear functions using the AMPL solver library \citep{gay1997hooking}, enables NN nonlinear activation functions. 
Second, Pyomo's many extensions, e.g., decomposition for large-scale problems, allow OMLT to interface with state-of-the-art optimization approaches.

Most importantly, OMLT uses Pyomo \emph{blocks}. In OMLT, Pyomo blocks encapsulate the GBT or NN components of a larger optimization formulation. Blocks simplify  OMLT: users only need to understand the input/output structure of the NN or GBT when linking to a Pyomo block's inputs and outputs.
The block abstraction allows OMLT users to ignore specific optimization details yet experiment with competing formulations. The block abstraction also helps OMLT developers wishing to create new optimization formulations and algorithms: Pyomo blocks provide flexible semantic structures for specialized algorithms.

\medskip

\noindent
\textbf{\texttt{OmltBlock} implementation}
\texttt{OmltBlock} is a Pyomo block that delegates generating the optimization formulation of the surrogate model to the \texttt{\_PyomoFormulation} object. OMLT users create the input/output objects, e.g., constraints, that link the surrogate model to the larger optimization problem and the user-defined variables. The optimization formulation of the surrogate is generated automatically from its higher level (ONNX or Keras) representation.
OMLT users may also specify a scaling object for the variables and a dictionary of variable bounds. The scaling and variable bound information may not be present in ONNX or Keras representations, but is required for some optimization formulations. Notebook \texttt{neural\_network\_formulations.ipynb} demonstrates using variable scaling and bounds.

\medskip

\noindent
\textbf{\texttt{NetworkDefinition} for neural networks}
For GBTs, OMLT automatically generates the optimization formulation from the higher level, e.g., ONNX, representation. For neural networks, OMLT instead generates an intermediate representation (\texttt{NetworkDefinition}) that acts as the gateway to several alternative mathematical optimization formulations.

\medskip

\noindent
\textbf{Alternative optimization formulations}
A major thread of research develops new optimization formulations for machine learning surrogates
\citep{fischetti2018deep,NEURIPS2018_29c0605a,singh2019beyond,anderson2020strong,tjandraatmadja2020,dathathri2020enabling}. OMLT uses Pyomo blocks to formulate surrogate models as an optimization objective or as constraints.
The \omltversion{} \texttt{GBTBigMFormulation} uses the \cite{mivsic2020optimization} and \cite{mistry2021mixed} formulation and thereby simplifies our GBT-based black-box optimization tool ENTMOOT \citep{thebelt2021entmoot,THEBELT2022118061}.
The OMLT NN implementation supports dense and convolutional layers.
For users developing custom formulations, we suggest \texttt{FullSpaceNNFormulation}.
OMLT also offers specific formulations which often arise in practice; we group these with respect to smooth versus non-smooth activation functions:
\begin{itemize}
    \item \textbf{Smooth}. \omltversion{} supports linear, softplus, sigmoid, and tanh NN activation functions with 2 competing formulations: \texttt{FullSpaceSmoothNNFormulation} and \texttt{Reduced} \texttt{SpaceSmoothNNFormulation} \citep{schweidtmann2019deterministic}.
    The full-space formulation uses Pyomo variables and constraints 
    to represent auxiliary variables and activation constraints. The reduced-space formulation instead uses Pyomo \texttt{Expression} objects to create a single constraint for each NN output.
    \item \textbf{Non-smooth}. \omltversion{} supports ReLU NN activation with competing formulations \texttt{ReluBigMFormulation} \citep{anderson2020strong}, \texttt{ReluComplementarityFormulation} \citep{yang2021modeling}, and \texttt{ReluPartitionFormulation} [dense layers only in \omltversion{}] \citep{tsay2021partition}.
\end{itemize}

Subclasses of \texttt{\_PyomoFormulation}, representing the full-space and reduced-space formulations, take an \texttt{OmltBlock} and allow us to represent the surrogate model using mathematical constraints and variables.
The full and reduced-space subclasses have \texttt{\_DEFAULT\_} \texttt{ACTIVATION\_CONSTRAINTS} for each activation function. Users wishing to try different optimization formulations beyond the provided alternatives should override these defaults.
An existing alternative for ReLU activation functions is the \texttt{ReLUPartitionFormulation}. Notebook \texttt{neural\_network\_formulations} experiments with competing formulations.

\section{Comparison to related work}

Subsets of \omltversion{} are available elsewhere.\footnote{\url{https://git.rwth-aachen.de/avt-svt/public/MeLOn}, \url{http://janos.opt-operations.com}, \url{https://github.com/ChemEngAI/ReLU\_ANN\_MILP}, \url{https://github.com/hwiberg/OptiCL}, \url{https://github.com/Gurobi/gurobi-machinelearning}} MeLOn \citep{schweidtmann2019deterministic} parses its own XML representation of dense NNs with sigmoidal activations and creates a reduced-space optimization formulation.
JANOS \citep{bergman2022janos} parses scikit-learn representations of dense ReLU NNs and logistic regression. Janos creates a Gurobi formulation to optimize over these surrogate models.
reluMIP \citep{reluMIP2021} parses TensorFlow representations of dense NNs with ReLU activation functions and creates a big-M formulation. OptiCL \citep{maragno2021mixed} creates its own machine learning surrogates and then develops mixed-integer formulations of its own surrogates. gurobi-machinelearning interfaces directly with the Gurobi solver.

OMLT is a more general tool incorporating both NNs and GBTs, many input models via ONNX interoperability, both dense and convolutional layers, several activation functions,\footnote{In \omltversion: ReLU, linear, softplus, sigmoid, and tanh} and various  optimization formulations. The literature often presents these different optimization formulations as competitors, e.g., our \emph{partition-based} formulation competes with the \emph{big-M} formulation for ReLU NNs \citep{kronqvist2021,tsay2021partition}. In OMLT, competing optimization formulations become alternatives: users can switch between the Section \ref{sec:library-design} formulations and find the best for a specific application.

\section{Outlook \& conclusions}

Higher-level representations, such as those available in ONNX, Keras, and PyTorch, are very useful for modelling neural networks and gradient-boosted trees. OMLT extends the usefulness of these representations to larger decision-making problems by automating the transformation of these pre-trained models into variables and constraints suitable for optimization solvers.
In other words, OMLT allows us to extend any optimization model that can be expressed in Pyomo to include neural networks or gradient-boosted trees.
Our implementation makes it possible to seamlessly compare different formulations that are typically presented as competitors in the literature.

\section{Acknowledgements}

FC, CT, and RM were funded by Engineering \& Physical Sciences Research Council Fellowships ( EP/T001577/1 \& EP/P016871/1). CT was also supported by a Imperial College Research Fellowship. BASF SE funded the PhD studentship of AT. \\[-6pt]

\noindent
{Sandia National Laboratories is a multimission laboratory managed and operated by National Technology and Engineering Solutions of Sandia, LLC, a wholly owned subsidiary of Honeywell International, Inc., for the U.S. Department of Energy’s National Nuclear Security Administration under contract DE-NA0003525. This paper describes objective technical results and analysis. Any subjective views or opinions that might be expressed in the paper do not necessarily represent the views of the U.S. Department of Energy or the United States Government. JJ and JH were funded in part by the Institute for the Design of Advanced Energy Systems (IDAES) with funding from the Office of Fossil Energy, Cross-Cutting Research, U.S. Department of Energy. JJ, JH, and CL were also funded by Sandia National Laboratories Laboratory Directed Research and Development (LDRD) program.}

\vskip 0.2in
\bibliography{omlt}

\end{document}